\documentclass[letterpaper, 10 pt, journal, twoside]{IEEEtran}

\usepackage[table]{xcolor}
\usepackage{mathtools,amssymb,commath}
\usepackage{amsmath}
\usepackage{graphicx,import}
\usepackage{verbatim}
\usepackage{color, soul}
\usepackage{lipsum}
\usepackage{verbatim}
\usepackage{algorithm}
\usepackage[noend]{algpseudocode}
\usepackage{subfig}
\usepackage{multicol}
\usepackage{tikz}
\usepackage{booktabs}
\usepackage[utf8]{inputenc}
\usepackage[english]{babel}
\usepackage{todonotes}
\usepackage{diagbox}
\usepackage{color, colortbl}
\usepackage{microtype}
\addto\captionsenglish{\renewcommand*{\tablename}{TABLE}}

\usepackage{url}
\usepackage{breakurl}
\usepackage[breaklinks, citecolor=black]{hyperref}

\definecolor{Gray}{gray}{0.9}
\newcolumntype{g}{>{\columncolor{Gray}}c}

\usetikzlibrary{shapes,arrows}
\hypersetup{colorlinks=true}

\newcommand\gabrisays[1]{\textcolor{blue}{\textbf{Gabri}: #1}} 
\newcommand\giusesays[1]{\textcolor{red}{\textbf{Giuse}: #1}} 

\usepackage[skins]{tcolorbox}
\usepackage{balance}
\newtcolorbox{myframe}[2][]{%
  enhanced,colback=white,colframe=black,coltitle=black,
  sharp corners,boxrule=0.4pt,left=0pt,right=0pt,top=0pt,bottom=0pt,
  fonttitle=\itshape,
  attach boxed title to top left={yshift=-0.3\baselineskip-0.4pt,xshift=2mm},
  boxed title style={tile,size=minimal,left=0.5mm,right=0.5mm,
  colback=white,before upper=\strut},
  title=#2,#1
}
\DeclareCaptionFont{mysize}{\fontsize{8.5}{9.6}\selectfont}
\usepackage{mathtools,amsthm}

\newtheorem{theorem}{Theorem}
\newtheorem{lem}[theorem]{Lemma}

\makeatletter
\def\BState{\State\hskip-\ALG@thistlm}
\makeatother

\title{Modeling, Identification and Control  of \\  Model Jet Engines for Jet Powered Robotics }

\author{Giuseppe L'Erario$^{1}$, Luca Fiorio$^{1}$, Gabriele Nava$^{1, \hspace{0.5 mm} 2}$, Fabio Bergonti$^{1}$, \\
Hosameldin 
Awadalla Omer 
Mohamed$^{1}$, Emilio Benenati$^{1}$, Silvio Traversaro$^{1}$ and Daniele Pucci$^{1}$
	\thanks{$^{1}$Dynamic Interaction Control research line of Fondazione Istituto Italiano di Tecnologia,
		via San Quirico 19, Genoa, Italy. 
		{Email addresses: \tt\small firstname.surname@iit.it}}
    \thanks{$^{2}$DIBRIS, University of Genova, Genoa, Italy.}%
}

\begin{document}

\maketitle
 \thispagestyle{empty}
 \pagestyle{empty}

\begin{abstract}

The paper contributes towards the modeling, identification, and control of model jet engines. We propose a nonlinear, second order model in order to capture the model jet engines governing dynamics. The  model structure is identified by applying sparse identification of nonlinear dynamics, and then the parameters of the model are found via gray-box identification procedures. Once the model has been identified, we approached the control of the model jet engine by designing two control laws. The first is based on the classical Feedback Linearization technique, while the second one on the Sliding Mode control method. The overall methodology has been verified by modeling, identifying  and controlling two model jet engines, i.e. P100-RX and P220-RXi developed by JetCat, which provide a maximum thrust of 100 N and 220 N, respectively. 

\end{abstract}


 \section{INTRODUCTION}
\label{introduction}

\IEEEPARstart{F}{lying} vehicles are not new to the Robotics community. Quadrotors, tail-sitters, and scale airplanes are only few examples of the large variety of existing flying robots~\cite{aerialRobots}. 
Yet, despite decades of research in the subject, the current propulsion systems for aerial robots is still a barrier for \emph{heavy} flying vehicles, which often need bulky and non efficient propellers~\cite{shim2005autonomous, DLR_quad, AirBus_VSR700,TechnicalAnalysisVTOLUAV}.
This paper takes the first step towards the modeling and control of model jet engines, which possess large thrust-to-weight ratios and use high energy density fuels. It thus paves the way to Jet Powered Robotics and to the design of a new generation of aerial vehicles.

Flying robots possess the capacity of aerial locomotion. When attempting at augmenting this capacity with a degree of manipulation or
terrestrial locomotion,
flying robots soon become heavier and more complex to control. For instance, manipulation and aerial locomotion have been combined by  \emph{Aerial Manipulation}~\cite{8299552}, a branch of robotics often exemplified by a quadrotor equipped with a robotic arm~\cite{6943038}. In this case, the robotic arm attached to the flying robot is often required to meet strict weight requirements, thus limiting the interaction capability of the aerial manipulator.

Aerial and terrestrial locomotion have also been combined into single robots. These robots exhibit multimodal locomotion capabilities~\cite{7994965,5334433,6631208,1748-3190-10-1-016005,6696526}, i.e. they can fly and move on the ground using contacts. Multimodal robots are more energetically efficient than classical aerial vehicles, but they share the same drawback: a very limited interaction ability with the environment. So, it is tempting to complement multimodal robots with a degree of manipulation, but the current electric propulsion system is still a barrier for such an augmentation.

Recently, attempts at combining aerial and bipedal terrestrial locomotion have also attracted the attention of the robotics community. LEg ON Aerial Robotic DrOne, or Leonardo, at the Caltech’s Center for Autonomous Systems and Technologies (CAST), combines two robotic legs with  propellers to improve balancing and agility~\cite{LeoCaltech}. Analogously, at Guangdong University of Technology’s School of Automation, researchers are developing a legged robot with ducted fans installed at its feet. The goal is to allow the robot to take larger steps \cite{Jet-HR1}. Yet, none of these robots is endowed with a degree of manipulation.\looseness=-1

\begin{figure}[t]
	\hspace*{-0.45cm}
	\centering
	\subfloat[][{}\label{fig:P100}]
	{\includegraphics[width=.20\textwidth]{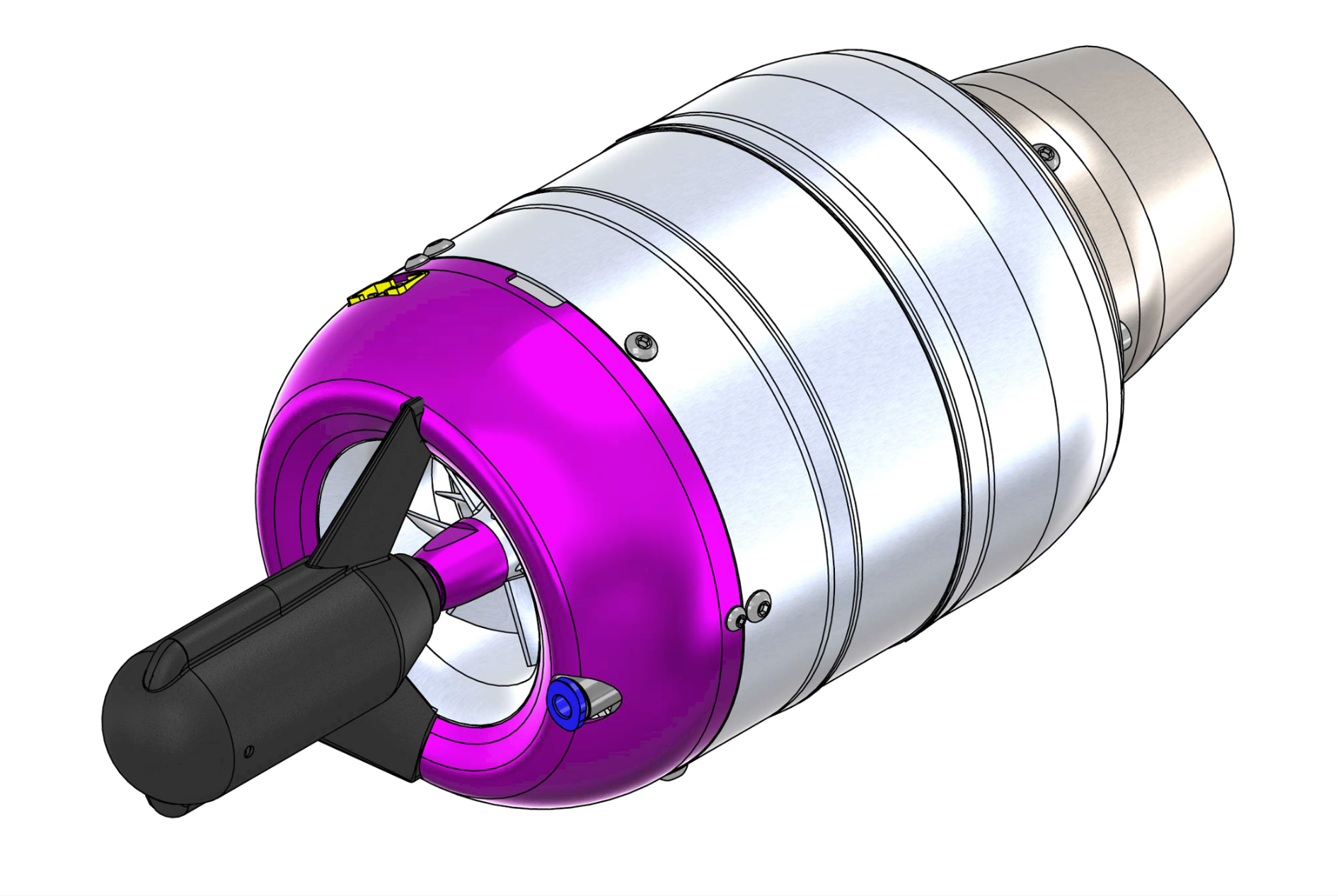}} 
	\subfloat[][{}\label{fig:P220}]
	{\includegraphics[width=0.34\textwidth]{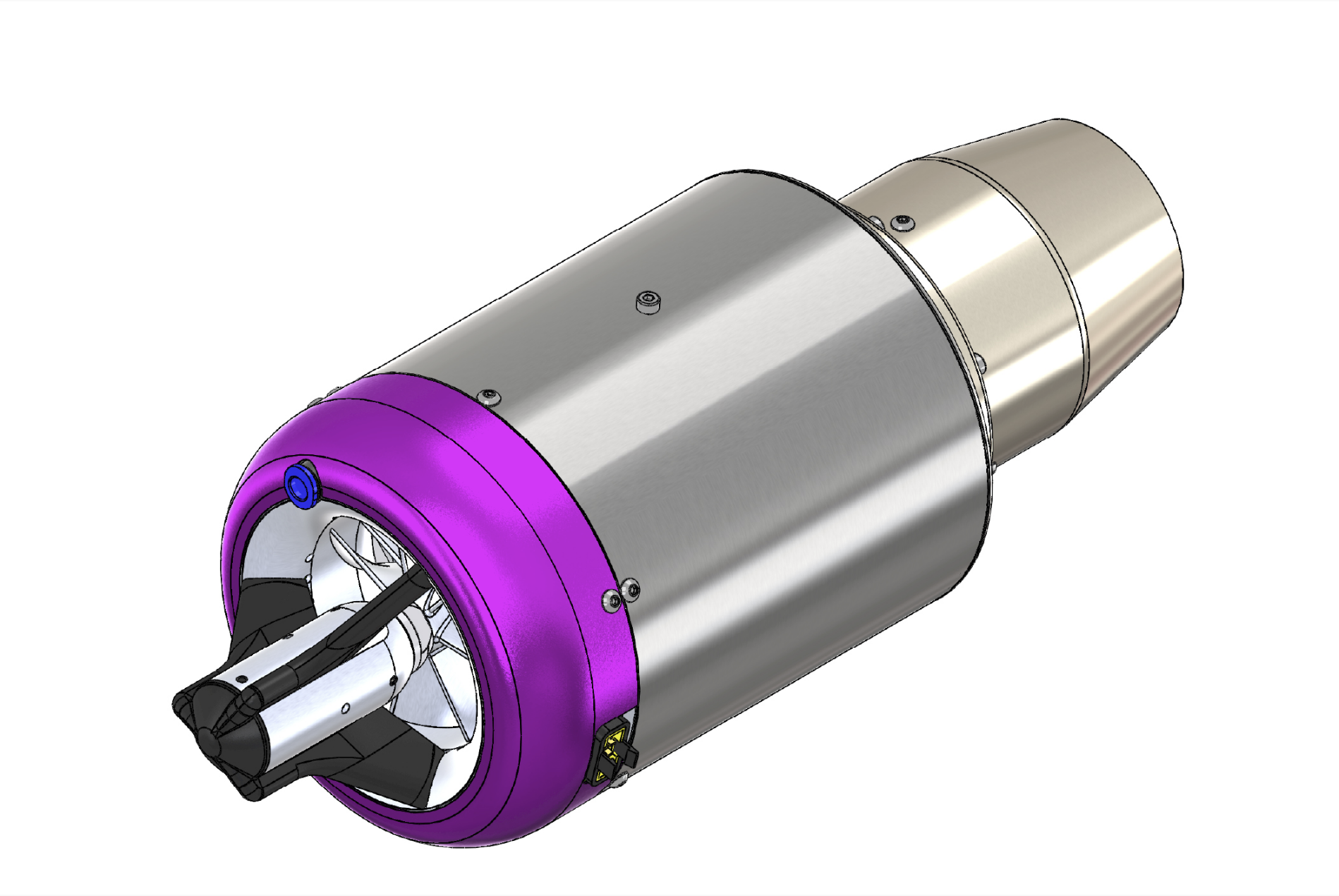}} \hspace{0.10cm}
	\caption{The model jet engines used for the experimental analysis: The P100-RX (a) and the P220-RXi (b) produced by JetCat.} \label{fig:jet_engine}
\end{figure}

An attempt at unifying manipulation, aerial, and bipedal terrestrial locomotion on a single robotic platform is given by the iRonCub project, whose aim is to make the humanoid robot iCub fly~\cite{Bartolozzi2017iCubTN, pucci2017momentum,nava2018position}. For such applications, where the propulsion system has to provide enough thrust to lift a complete humanoid robot, electric ducted fans are no longer suitable. An option is then to choose high thrust-to-weight ratio propulsion systems, as \emph{model jet engines} -- see Table~\ref{fig:jet_engine}.

\begin{figure*}[ht]
 \vspace*{0.4cm}
 \captionof{table}{The weight in kg of the battery bank in case of electric propulsion (a). The weight in kg of fuel in case of jet propulsion (b).}
 \centering
 \begin{myframe}{Electric VS Jet propulsion}
 	\vspace{0.05cm}
 	\hspace{1.0cm}
 	\subfloat[][{Electric propulsion.}\label{fig:electric_prop}]
     	{\begin{tabular}{c|c|c|c}
        \toprule
        \diagbox[width=10em]{Robot \\ weight [kg]}{Flight time \\ $[\text{min}]$} & 1 & 3 & 5  \\ 
        \midrule
        \rowcolor{gray!15}
        10 & 0.86   &  3.12  &   6.57   \\ 
        20 & 1.72  &  6.25  &  13.15   \\
        \rowcolor{gray!15}
        30 & 2.58   &  9.37  &  19.73   \\
        40 & 3.44   & 12.50  & \cellcolor{red!40} 26.31   \\
        \rowcolor{gray!15}
        50 & 4.31   & 15.62  &  32.89   \\
        \bottomrule
        \end{tabular}} \quad \hspace{1.2cm}
 	\subfloat[][{Jet propulsion.}\label{fig:jet_prop}]
     	{\begin{tabular}{c|c|c|c}
        \toprule
        \diagbox[width=10em]{Robot \\ weight [kg]}{Flight time \\ $[\text{min}]$} & 1 & 3 & 5  \\ 
        \midrule
        \rowcolor{gray!15}
        10 & 0.22  &  0.67  &  1.15  \\
        20 & 0.44  &  1.35  &  2.30  \\
        \rowcolor{gray!15}
        30 & 0.66  &  2.02  &  3.45  \\
        40 & 0.88  &  2.70  &  \cellcolor{red!40} 4.61  \\
        \rowcolor{gray!15}
        50 & 1.10  &  3.38  &  5.76  \\
        \bottomrule
        \end{tabular}}
    \end{myframe}
 \label{tab:elec_VS_jet}
 \vspace{-0.0cm}
\end{figure*}

Despite the vast literature on aircraft jet engines (see, e.g.,~\cite{hall2013fluid} and the references therein), a comprehensive reference for the modeling, identification, and control of model jet engines is still missing. Consequently, the application of model jet engines to Robotics remains  at an embryonic stage. To get started with this application, one may be tempted to look at the several textbooks on jet engine design, which  provide precise models of the underlying  engine. Thus, one would analyse the turbine working cycle, internal systems,  performance, and  maintenance~\cite{jetEnginesRollsRoice}, and would finally obtain a model of the model jet turbine. This model, however, depends on several variables such as the internal geometry, the fuel heating value, the use condition, and the air density: all these information and measurements are seldom available on classical robot sensor suits.
Therefore, alternative  methods shall be developed to circumvent the modelling and control challenges given by model jest turbines. \looseness=-1

This paper takes the first step to enable the use of model jet engines in robotics applications. More precisely, the manuscript presents modelling, identification, and control techniques for model jet engines, and performs experimental validation of the proposed  methods. To the best of the authors knowledge, the paper is the first comprehensive treatment of model jet engines, ranging from modeling to experimental validation. 
Starting from experimental data, we first use the sparse identification method SINDy~\cite{Brunton3932} to find out the governing jet engine model. After fixing the model structure, 
we perform 
\textit{gray-box} estimation using two methods: batch least squares and recursive extended Kalman Filter. The overall modelling and identification procedure points out the jet turbine model, which is then used in the control design stage. 
We tackle the control of the model jet engines by applying and comparing \textit{feedback linearization} and \textit{sliding mode} control~\cite{isidori2013nonlinear,khalil2002nonlinear}. The overall approach presented in the paper is verified experimentally on the model jet engines P100-RX and P220-RXi developed by JetCat. These experimental activities are carried out in a dedicated setup that has been designed and built for the purpose of this manuscript.\looseness=-1

The paper is organized as follows. Section~\ref{why} motivates the use of model jet engines in Robotics. Section~\ref{sec:background} recalls the extended Kalman filter algorithm and Sliding Mode control. Section~\ref{sec:testBench} describes the test bench and the hardware used for the experimental analysis.  Section~\ref{sec:method_estimation} describes the identification procedure with also some validation results. Section~\ref{sec:method_control} presents  a control architecture for model jet engines and the associated experimental results. Section~\ref{sec:conclusion} closes the paper with conclusions and perspectives.

\section{WHY JET POWERED ROBOTICS}
\label{why}

Electric propulsion systems are nowadays the common actuators for flying robots. The relative simplicity in using brushless and brushed motors, and the advent of efficient and strong propellers paved the way to  flying robots powered by electric propulsion. One of the drawbacks of electric powered flying robots is clearly the need of high power, and consequently large capacity batteries, when the robot weight increases considerably. In these cases, model jet engines, which produce large thrust-to-weight ratios using high energy density fuels, may be a solution for \emph{heavy} flying robots.

For the purpose of understanding when model jet engines should be preferred to  electrical motors, we consider the problem of making a robot -- of a given weight --  hover for a specific amount of time. Then, we  estimate the associated weight of LiPo battery banks and jet fuel in the electrical and jet engine case, respectively. We assume that the specific energy density  of the LiPo batteries is $210 \ \text{Wh/kg}$ \cite{LIPO_ulvestad2018brief, LIPO_endurance}, and that the density of the jet fuel (i.e. diesel) is $0.8 \ \text{kg/l}$ \cite{marketing2007diesel}.

More specifically, the analysis is carried out considering the performances of the following two propulsion systems:
\begin{itemize}
    \item \textbf{Dynamax CAT 6 FAN} \cite{Cat6Fan}, an electric motor with a maximum thrust of $\sim 13 \ \text{kg}$ at $\sim 13 \ \text{kW}$ consumption;
    \item \textbf{JetCat P220-RXi} \cite{P220_ref}, a model jet engine with maximum thrust of $\sim 22 \ \text{kg}$ at $\sim 0.6 \ \text{l/min}$ consumption;
\end{itemize}
We also assume that energy and fuel consumption are linear versus the generated thrust, and that the engine weights are neglectable. The Table~\ref{tab:elec_VS_jet} shows the difference in terms of weight that is needed to power the robot for the electric and jet engine case, respectively. If we consider, for instance, a 40 Kg robot performing five minute flight, the advantage of using model jet engines is evident: we would need 26.31 Kg of battery pack versus 4.61 Kg of fuel. This specific considered case is that of the envisaged flying version of the humanoid robot iCub.

Another interesting element of comparison is the number of engines needed to lift a specific weight. Table~\ref{tab:engines_num} shows this comparison, and we observe that a 40 Kg robot performing hovering needs 4 electrical motors instead only 2 jet turbines. 

\begin{table}[t]
\vspace*{0.4cm}
	\centering
	\caption{Number of engines needed for hovering.}
    \begin{tabular}{c|c|c|c|c|c}
    \toprule
    Robot weight [kg] & 10 & 20 & 30 & 40 & 50  \\ 
    \midrule
    Electrical prop.  &  1  &  2  &  3  &  \cellcolor{red!40} 4  &  4  \\
    Jet prop.         &  1  &  1  &  2  &  \cellcolor{red!40} 2  &  3  \\
    \bottomrule
    \end{tabular}
    \label{tab:engines_num}
\end{table}

\section{BACKGROUND}
\label{sec:background}

\subsection{Notation}
\begin{itemize}
    \item $\mathbf{a} \in \mathbb{R}^i$ is a vector of dimension $i$;
    \item $\mathbf{A} \in \mathbb{R}^{i \times j}$ is a matrix of size $i$ times $j$;
	\item $\mathbf{x} \in \mathbb{R}^n$ is the state vector of a generic system;
	\item $\mathbf{u} \in \mathbb{R}^p$ is the input vector of a generic system;
	\item $\mathbf{y} \in \mathbb{R}^m $ is the output vector of a generic system;
	\item $\mathbf{z} \in \mathbb{R}^m $ is the measurement vector;
	\item $T \in \mathbb{R}^1$ is the thrust of the jet engine model; 
	\item $[T, \dot{T}]^\top \in \mathbb{R}^2 $ is the state vector of the jet engine model;
	\item $u \in \mathbb{R}^1$ is the input of the jet engine, namely the \textit{PWM}. 
\end{itemize}

\subsection{Recall on Kalman Filter}
We here recall elements of  the 
discrete Extended Kalman filter~\cite[Sec. 13.2.3]{simon2006optimal}.
\setcounter{equation}{1}
Consider a discrete nonlinear system:
\begin{IEEEeqnarray}{LL}
    \label{nonlinearSysteDiscr}
    \mathbf{x}_{k+1} &= f(\mathbf{x}_k, \mathbf{u}_k) + \mathbf{w}_k  \IEEEyessubnumber\\
    \mathbf{y}_k     &= h(\mathbf{x}_k)               + \mathbf{v}_k \IEEEyessubnumber
\end{IEEEeqnarray}
subject to process noise $\mathbf{w}_k$ and measurement noise $\mathbf{v}_k$. The noises are uncorrelated, zero mean Gaussian noises and have known covariance matrices $\mathbf{Q}_k {\in} \mathbb{R}^{n \times n}$ and $\mathbf{R}_k {\in} \mathbb{R}^{m \times m}$ describing the process and the measurement noise, respectively.

The covariance of the estimation error $\mathbf{P}_k$ is defined as:
\begin{equation}
	\mathbf{P}_k = E[(\mathbf{x}_k - \hat{\mathbf{x}}_k)(\mathbf{x}_k - \hat{\mathbf{x}}_k)^\top],
\end{equation}
with $\mathbf{P}_k \in \mathbb{R}^{n \times n}$, $\mathbf{x}_k$ and $\hat{\mathbf{x}}_k$  the true state and its estimation, and $E(\cdot)$ the operator denoting the expected value. 

The Extended Kalman Filter uses Taylor series to expand the nonlinear system~\eqref{nonlinearSysteDiscr} around the current state estimate and can be used to estimate the state of the system from a series of noisy measurements~$\mathbf{z}$.

\subsection{Recall on Sliding Mode controller}
\label{SM_theory}

Sliding Mode Control is a  robust control technique that 
forces the system trajectories to converge about a  manifold, where the system evolution is independent of the model uncertainties~\cite[p. 552]{khalil2002nonlinear}. This manifold is called \textit{sliding manifold}.
A drawback of Sliding mode controllers is the so called \textit{chattering}, which may excite unmodeled high-frequency dynamics that may degrade the  system performance. Next lemma recalls a Sliding mode controller combined with a common technique used to reduce chattering. \looseness=-1 

\begin{lem}
\cite[Sec. 14.1]{khalil2002nonlinear} Consider the 
system:
\begin{IEEEeqnarray}{LL}
	\dot{x}_1 & = x_2 \IEEEyessubnumber \\
	\dot{x}_2 & = f(\mathbf{x}) + g(\mathbf{x}) u, \IEEEyessubnumber
	\label{nonLinearSystem}
\end{IEEEeqnarray}
where $\mathbf{x}:=[x_1, x_2]$,  $f(\mathbf{x})$ and $g(\mathbf{x})$ are \textit{uncertain} nonlinear functions and $g(\mathbf{x}) > 0, \ \forall \mathbf{x} \in  \mathbb{R}^2$.

\noindent
	Assume that:
	\begin{itemize}
		\item $\hat{f}(\mathbf{x})$ and $\hat{g}(\mathbf{x})$ represent the nominal model;
		\item $\beta(\mathbf{x}) \ge \Big|\dfrac{a_1 x_2 + f(\mathbf{x})}{g(\mathbf{x})}\Big| + \beta_0$, $\beta_0 >0$. 
	\end{itemize}
	Then, the control law:
	\begin{equation}
	u = -\underbrace{\frac{a_1 x_2 + \hat{f}(\mathbf{x})}{\hat{g}(\mathbf{x})}}_{\textnormal{continuous component}} - \underbrace{\beta(\mathbf{x}) \cdot \textnormal{sgn}(s)}_{\textnormal{switching component}}
	\label{sliding_controller}
	\end{equation}
	ensures that the system trajectories converge to
	\begin{equation*}
	    s = a_1 x_1 + x_2 = 0.
	\end{equation*}
\end{lem}
\noindent
On the sliding manifold $s = 0$, the motion is governed by $\dot{x}_1 = - a_1 x_1$. If $a_1 > 0$, then $\mathbf{x}(t)$ tends to zero, and the constant $a_1$  regulates the rate of convergence towards zero. 

\begin{figure}[t]
	\centering
	\includegraphics[width=1\linewidth]{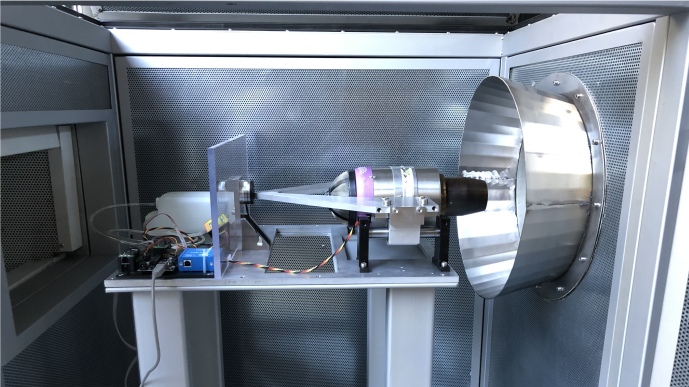}
	\caption{The setup used for jet experimental campaigns.}
	\label{fig:test_setup}
\end{figure}


\section{THE TEST BENCH \\ FOR JET EXPERIMENTAL ACTIVITIES}
\label{sec:testBench}

Carrying out experimental activities with jet turbines is not a straightforward activity. Exhaustion gas at about 800$^\circ$ Celsius, flammable fuel, and rotating turbines at about 200000 RPMs 
are only few challenges that, if not dealt with carefully, may turn to be hazardous for the staff conducting the experimental campaigns. 
In this section, we describe the setup for jet turbines 
that was designed and built to secure the experimental activities presented in this paper.  

First, the setup had to be compatible with two types of model jet engines: the  P100-RX and the P220-RXi by JetCat~\cite{P100_ref,P220_ref}. Each turbine comes with a dedicated Electronic Control Unit (ECU) and a Ground Support Unit (GSU). 
The ECU can work through two different communication interfaces: throttle and serial. The throttle interface is the most basic one, and accepts a Pulse With Modulation (PWM) input signal to regulate the turbine thrust. The serial interface receives the desired throttle as a message, but can also send a feedback message that details the running status of the engine.
The default engine fuel is kerosene Jet-A1, but the performances are similar also if diesel is used.
The main characteristics of the engines are in Table~\ref{tab:turbines_specs}.


\begin{table}[b]
\caption{Basic specifics of the jet engines.}
	\centering
	\begin{tabular}{c|c|c}
		\toprule
		Jet engine model     & P100-RX & P220-RXi \\
		\midrule
		\rowcolor{gray!15}
		Nominal Max. Thrust  & $100$ N & $220$ N \\
		Throttle range & $25 \% \ - \ 100 \% $ & $25 \% \ - \ 100 \% $ \\
		\rowcolor{gray!15}
		Weight               & $1080$ g & $1850$ g \\
		Length               & $241$ mm & $307$ mm \\
		\rowcolor{gray!15}
		Diameter             & $97$ mm & $116.8$ mm \\
		\bottomrule
	\end{tabular}
	\label{tab:turbines_specs}
\end{table}


Due to their dangerous nature, jet turbine experimental activities required the design and construction of a dedicated protecting case.
Fig.~\ref{fig:test_setup} depicts the interior of the turbine case. 
On the left side, we see the air intake opening; at the center, the turbine instrumentation; on the right, the exhaust cone. The case has been designed to be explosion and fire proof.

\begin{figure}[t]
	\centering
	\includegraphics[width=1\linewidth]{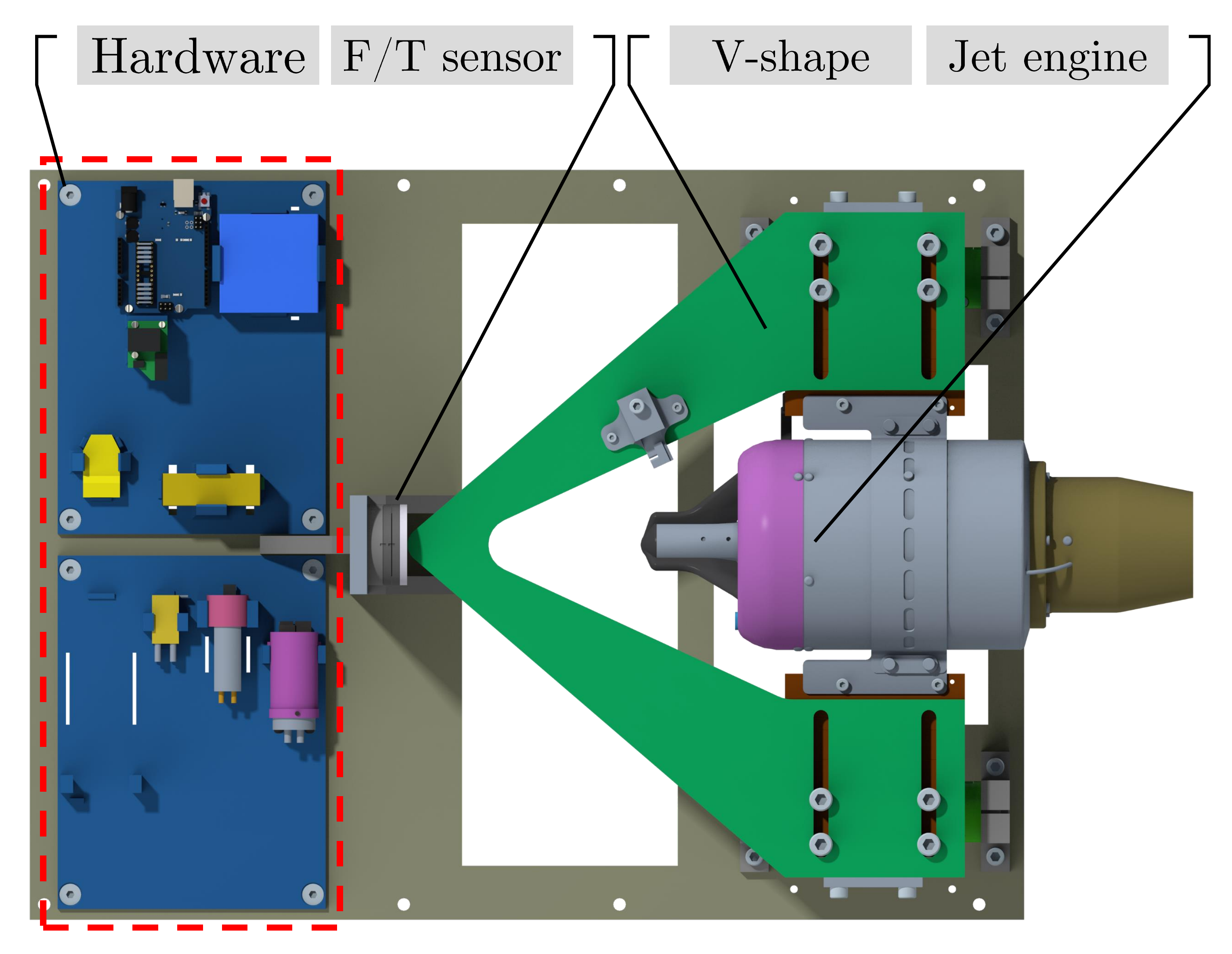}
	\caption{
	The jet engine instrumentation. Inside the dashed red box, the electronics and pump supports; in green, the turbine ``V-shape" support.}
	\label{fig:setup}
\end{figure}

As depicted in Fig. \ref{fig:setup}, the turbine instrumentation consists in a custom designed calibration device and the supports for electronics.
More precisely, the calibration device features a V-shaped frame, supported by linear bearings, to which the jet engines are anchored. 
The turbine thrust is measured by an IIT custom made force-torque sensor (F/T) 
in contact with the V-shaped frame. 
For what concerns the electronics, we used an ``Arduino-like" 
board to generate the PWM signal for the throttle interface of the ECU and a USB-CAN box to read the force measurements from the F/T.\looseness=-1




Both the board and the box were connected through a USB port to a laptop running YARP \cite{ceseracciu2013middle} and a MATLAB/Simulink controller. 
The electronics communicates thanks to YARP devices and ports, while the communication between YARP and the controller was implemented through the Whole-Body Toolbox \cite{RomanoWBI17Journal,ferigo2019generic}.
The whole control architecture runs at a frequency of 100 Hz.



\section{MODELING AND IDENTIFICATION \\ OF MODEL JET TURBINES}
\label{sec:method_estimation}



\begin{figure}[t]
\centering
 	\subfloat[][{Steps.}\label{fig:datasetp100_steps}]
 	{\includegraphics[width=0.48\textwidth]{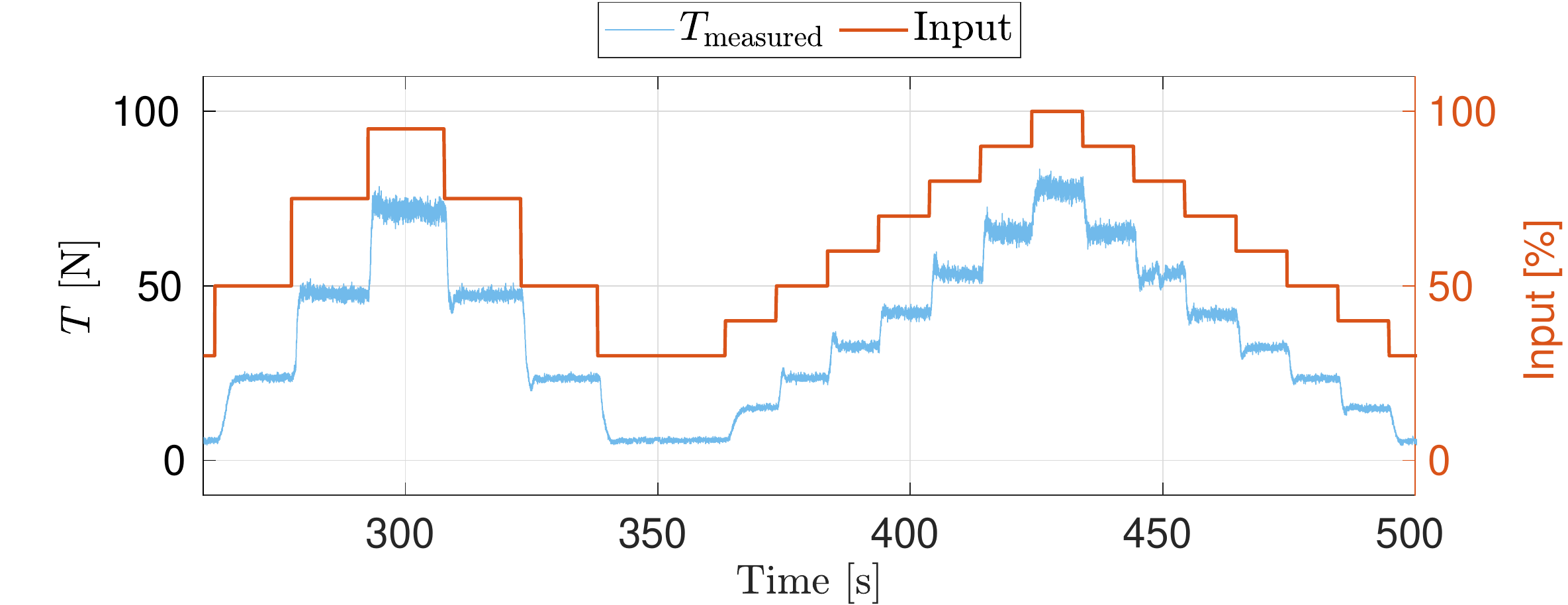}} \\
 	\subfloat[][{Chirps with different amplitude and equilibrium  points.}\label{fig:datasetp100_chirps}]
	{\includegraphics[width=0.48\textwidth]{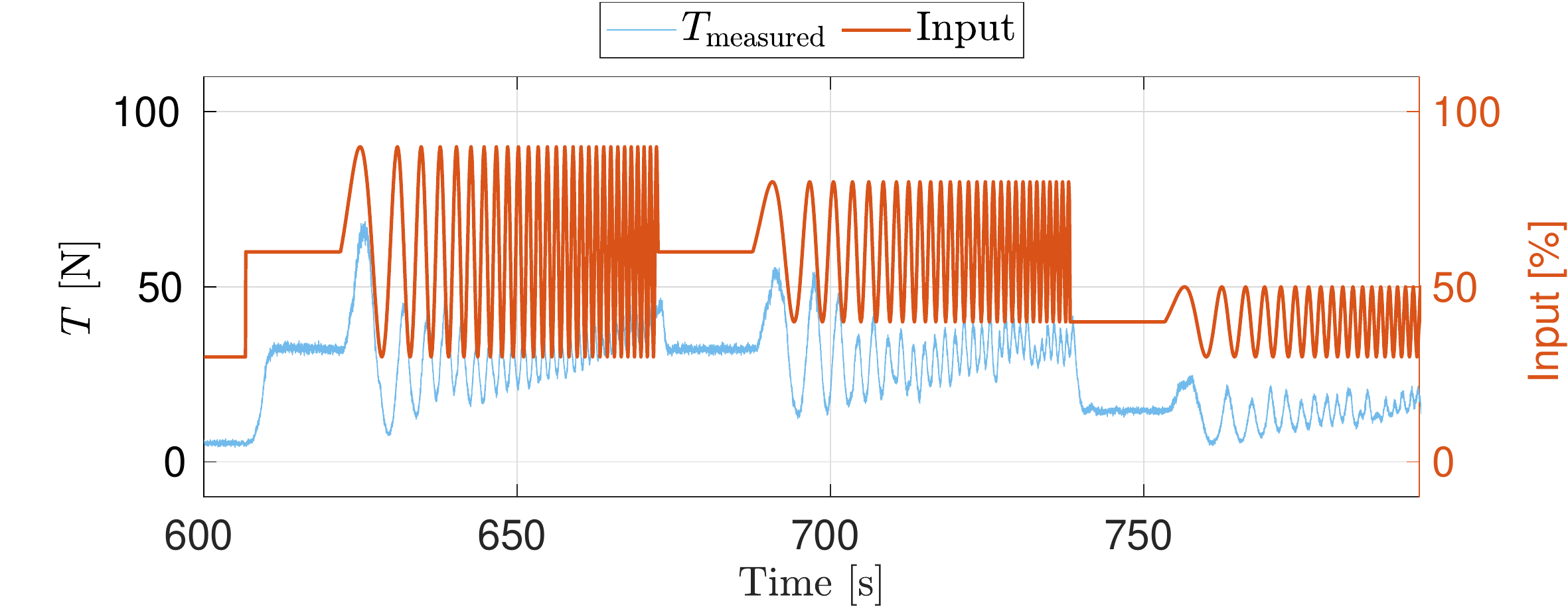}}
 	\caption{Examples of  experimental dataset with the P100-RX  jet engine.} \label{fig:datasetp100}
\end{figure}

The test bench presented in Sec.~\ref{sec:testBench} is used to collect input (i.e. throttle) and output (i.e. thrust) data. An example of dataset collected during an experimental campaign is depicted in Fig.~\ref{fig:datasetp100}. 
From a qualitative 
investigation, it is clear that second order systems for the model jet engine are good initial candidates for performing \emph{gray-box} identification. 

\subsection{Identification of jet model structure}
\label{structure}

Following the observation that second order models may be good initial candidates,
we apply a sparse identification method for nonlinear dynamics called SINDy~\cite{Brunton3932}. SINDy is a state-of-the-art technique to find  functions describing the relationships between variables and measured dynamics. 
In particular, from an augmented library matrix of $n_p$ arbitrary candidate functions, SINDy selects the governing terms among these functions using a sparse feature selection mechanism, such as LASSO regression or the sequential thresholded least-squares algorithm.
To apply SINDy for model jet engine identification, assume that the thrust dynamics is given by:
\begin{equation}
    \ddot{T} = f(T, \dot{T},u),
\end{equation}
where the thrust $T$ is the only available measurement, and $\dot{T}$ and $\ddot{T}$ are retrieved using the Savitzky–Golay filtering~\cite{schafer2011savitzky}. 
The SINDy model library -- that is represented internally by a matrix -- is composed of a combination of the state $[T, \dot{T}]$ and input $u$ for a total of $n_p~=~56$ terms. Furthermore, every dataset is composed of at least $n_s = 30000$~samples. Let us  remark that to run the SINDy sparse identification, we used datasets with an input similar to that of Fig.~\ref{fig:datasetp100}. Namely, it is composed of steps, ramps, sinusoids and chirps that induce a thrust force. The chosen dataset for the input-output is a sufficiently ``exciting'' system trajectory since it guarantees that the regression matrix used inside SINDy is of full rank, and thus the identification result is meaningful.

The application of the sparse optimization
yields~\cite{Brunton3932} :
\begin{IEEEeqnarray}{LL} 
    \ddot{T}  = & a_1 + a_2 T + a_2 T^2 + a_3 \dot{T} + a_4 T \dot{T} + a_5 \dot{T}^2 \nonumber \\
    & + a_6 u + a_7 T u + a_8 \dot{T} u + a_9 u^2,
\end{IEEEeqnarray}
with also an estimate of the coefficients $a_i$. 
Let us observe that simulations we have performed tend to show that the terms depending on the control input $u$, namely, $(a_6 \cdots a_9)$, are very sensitive to the tuning parameters of SINDy.
This  suggests that the nonlinearities depending on the control input $u$ do not play a pivotal role for identification purposes. Then, we decide to fix the model shape 
so as the resulting nonlinear dynamics is affine versus the  input~$u$, i.e.
\begin{equation}
	\ddot{T} = f(T,\dot{T}) + g(T, \dot{T}) \ v(u),
	\label{turbine_model}
\end{equation}
where 
\begin{IEEEeqnarray*}{LL}
f(T, \dot{T}) & = K_T T {+} K_{TT} T^2 {+} K_D \dot{T} {+} K_{DD}\dot{T}^2 {+} K_{TD} T \dot{T} + c, \\
	 g(T, \dot{T}) & = B_U + B_T T + B_D \dot{T}, \\
	 v(u) & = u + B_{UU}u^2,
\end{IEEEeqnarray*}
and $u$ the jet engine input. The model~\eqref{turbine_model} is particularly handy 
for  identification and control purposes presented next.

\subsection{Gray-box identification of the jet model}\label{EKF_grayIdentification}
\label{gray-box}

%
%
%
%
%
%

Once the model~\eqref{turbine_model} is fixed, we apply gray box identification to estimate its parameters. Define the parameter vector:
\begin{equation*}
	\mathbf{p}{=}(K_T, K_{TT},  K_D,  K_{DD},  K_{TD},  c, B_U, B_T,  B_D,  B_{UU})^\top,
	\label{param_vect}
\end{equation*}
 where every element of the vector is a scalar.
Then, the goal of the gray box identification is to estimate the vector $\mathbf{p}$.

The model~\eqref{turbine_model} may be expressed linearly versus $\mathbf{p}$ when linearized. 
So, one may attempt at identifying the set of parameters by performing a set of experimental activities from which $\dot{T},\ddot{T}$ are estimated via non causal filters (e.g.  Savitzky–Golay) and the set of parameters $\mathbf{p}$ extracted using classical iterative linear regression. The application of this method -- whose  results are shown in the section next -- turned to be not robust with respect to measurement noise. 



Another approach to parameter estimation is that of including the system parameters in an Extended Kalman Filter (EKF) that iteratively estimates the vector $\mathbf{p}$. More precisely, we can define a system so as  the state $\mathbf{x}$ and its parameters $\mathbf{p}$ are contained in the system dynamics. 
The augmented state vector $\mathbf{x}'$ is then defined as:
\begin{equation}
\begin{split}
\mathbf{x}'_{k+1} = 
\begin{bmatrix}
\mathbf{x}_{k+1} \\
\mathbf{p}_{k+1}
\end{bmatrix}  & =
\begin{bmatrix}
f(\mathbf{x}_k) + g(\mathbf{x}_k) \mathbf{v}_k + \mathbf{w}_k \\
\mathbf{p}_k + \mathbf{w}_{pk}
\end{bmatrix} = \\
& = f'(\mathbf{x}'_k, \mathbf{v}_k, \mathbf{w}_k, \mathbf{w}_{pk}),
\end{split}
\end{equation}
being $\mathbf{w}_{pk}$ an artificial noise 
that allows the EKF to change the estimate of $\mathbf{p}_k$ and $f(\mathbf{x}'_k, \mathbf{v}_k, \mathbf{w}_k, \mathbf{w}_{pk})$ a nonlinear function of the augmented state $\mathbf{x}'_k$.

To apply the above EKF based method to jet engine parameter identification, we first discretize~\eqref{turbine_model} and then  extend the state with the parameter vector $\mathbf{p}$, namely:
\begin{equation}
\begin{split}
\mathbf{x}'_{k+1}&=
\begin{bmatrix}
	T_{k+1} \\
	\dot{T}_{k+1} \\ 
	\mathbf{p}_{k+1} 
\end{bmatrix} = \begin{bmatrix}
    T_{k} + \dot{T}_{k}\Delta t + \ddot{T}_k\frac{\Delta t^2}{2} \\
    \dot{T}_{k} + \ddot{T}_k \Delta t  \\
    \mathbf{p}_{k} 
\end{bmatrix} \\
y_{k} &= T_k,
\end{split}
\end{equation}
where $\ddot{T}_{k}$ is the second derivative \eqref{turbine_model} of the thrust evaluated at time instant $k$ and $\Delta t$ the sampling interval. 

The EKF is then used to estimate the  state $[T, \dot{T}, \mathbf{p}]^\top$.
The  estimation is  corrected using the only available measurement, i.e. the thrust $T$, which is independent of its numerical derivatives $\dot{T}, \ddot{T}$.
Since we are interested  in estimating $\mathbf{p}$, we run  procedure offline and on the collected datasets. 
The choices of the initial state guess,  covariance error, and noise matrices are critical for the EKF tuning. In our  case, the initial guess of the parameters $\mathbf{p}$ influences considerably the overall results associated with the identification of $\mathbf{p}$ itself.  We solved this issue by running iteratively the EKF algorithm on the dataset to improve, run by run, the initial guess, while performing a proper tuning of the matrices $\mathbf{Q}$, $\mathbf{R}$ and $\mathbf{P}$. 


\begin{figure*}[ht]
 \vspace*{0.4cm}
 \centering
 \begin{myframe}{Identification of the P100-RX}
 	\vspace{-0.2cm}
 	\hspace{0.5cm}
 	\subfloat[][{Calibration.}\label{fig:calibrationP100}]
 	{\includegraphics[width=.45\textwidth]{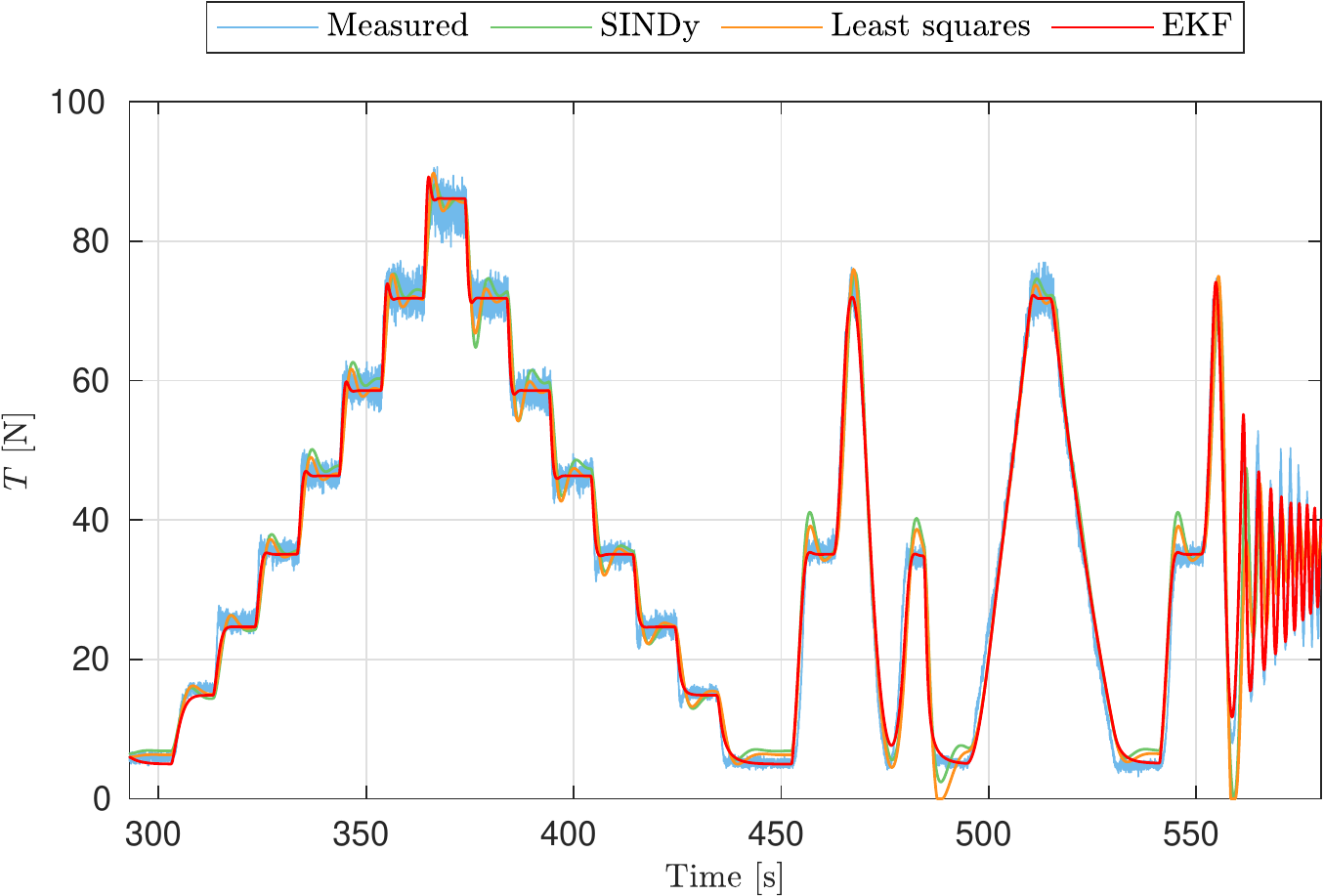}} \quad
 	\subfloat[][{Validation.}\label{fig:validationP100}]
 	{\includegraphics[width=.45\textwidth]{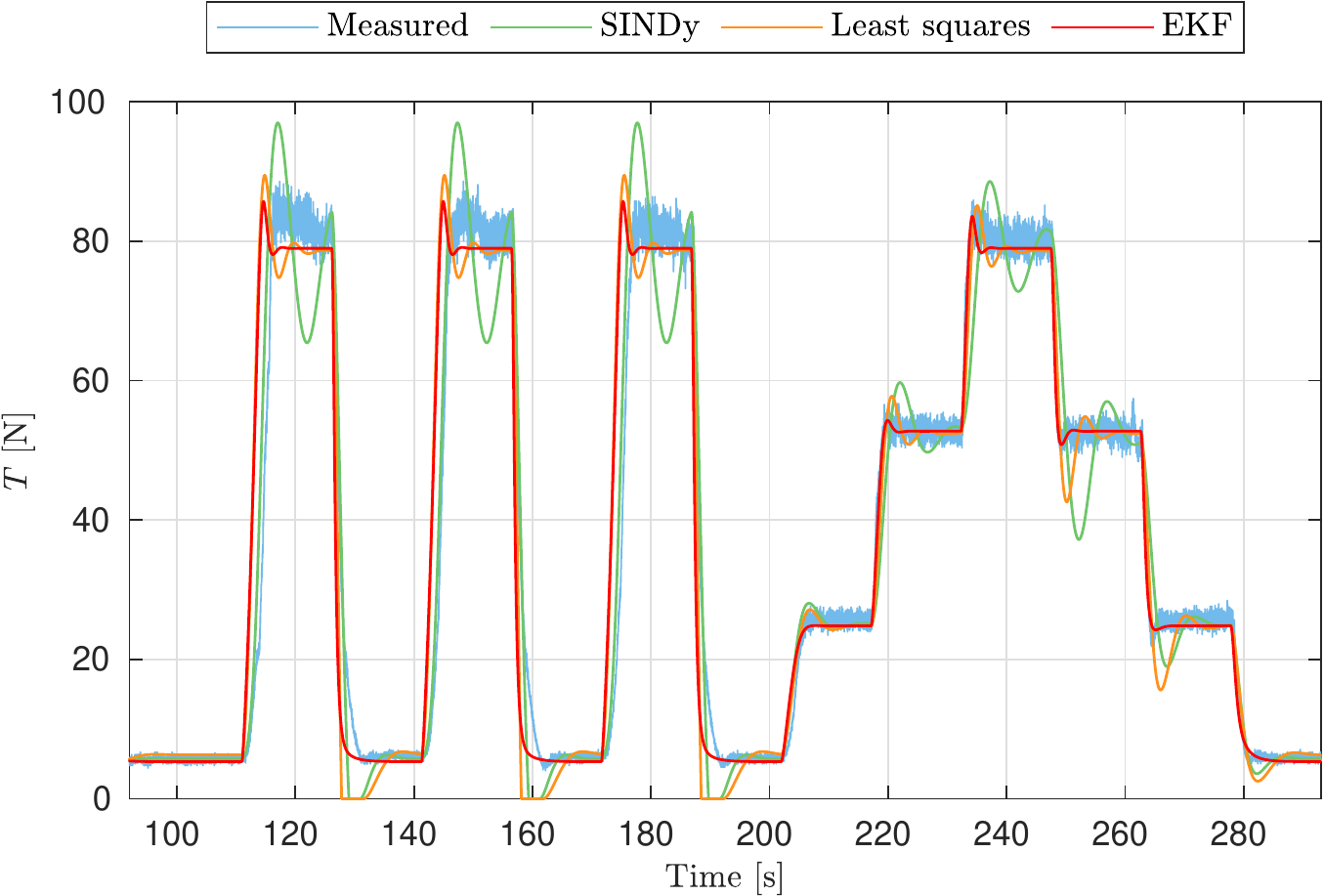}}
 \end{myframe}
 \centering
 \begin{myframe}{Identification of the P220-RXi}
 	\vspace{-0.2cm}
 	\hspace{0.5cm}
 	\subfloat[][{Calibration.}\label{fig:calibrationP220}]
 	{\includegraphics[width=.45\textwidth]{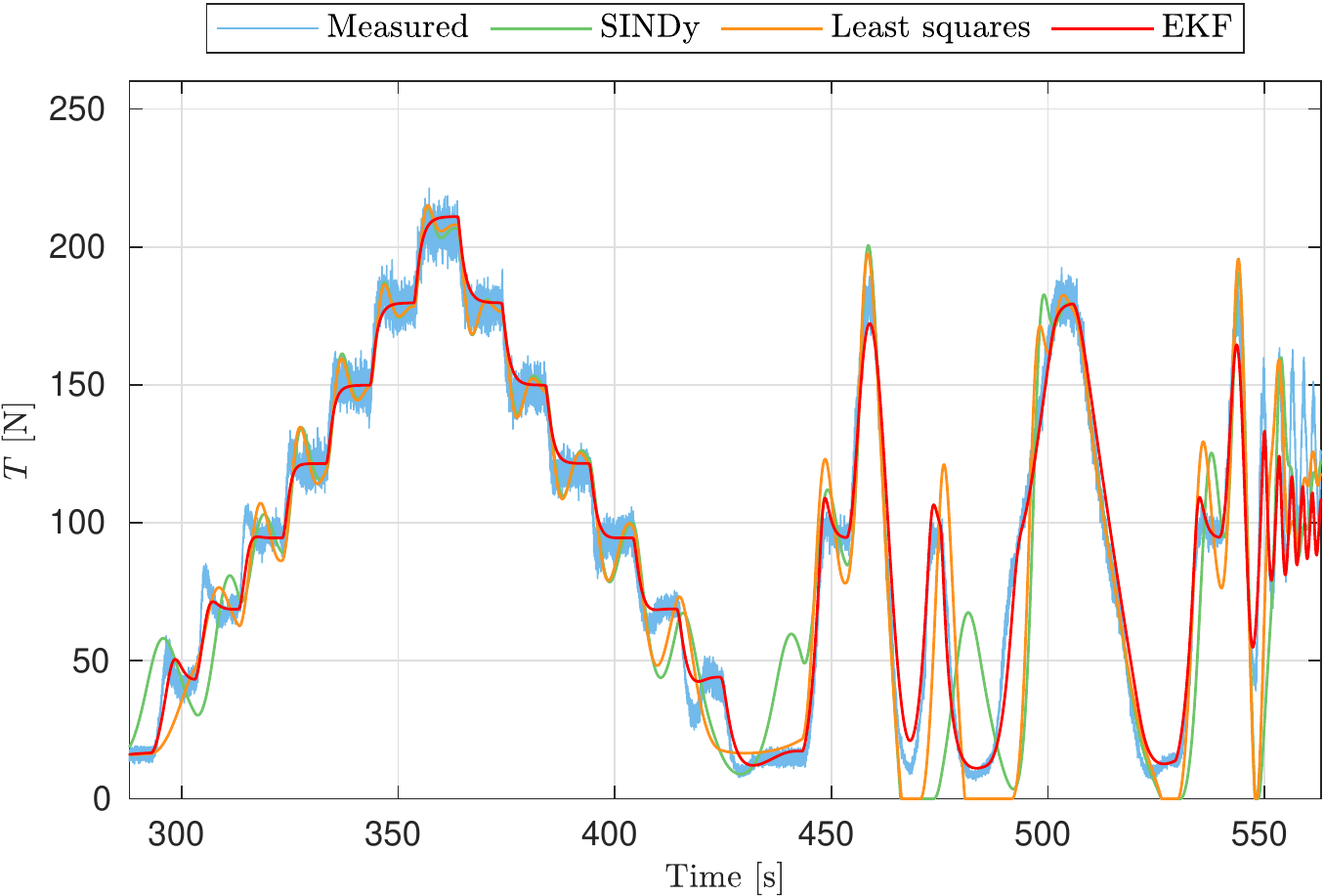}} \quad
 	\subfloat[][{Validation.}\label{fig:validationP220}]
 	{\includegraphics[width=.45\textwidth]{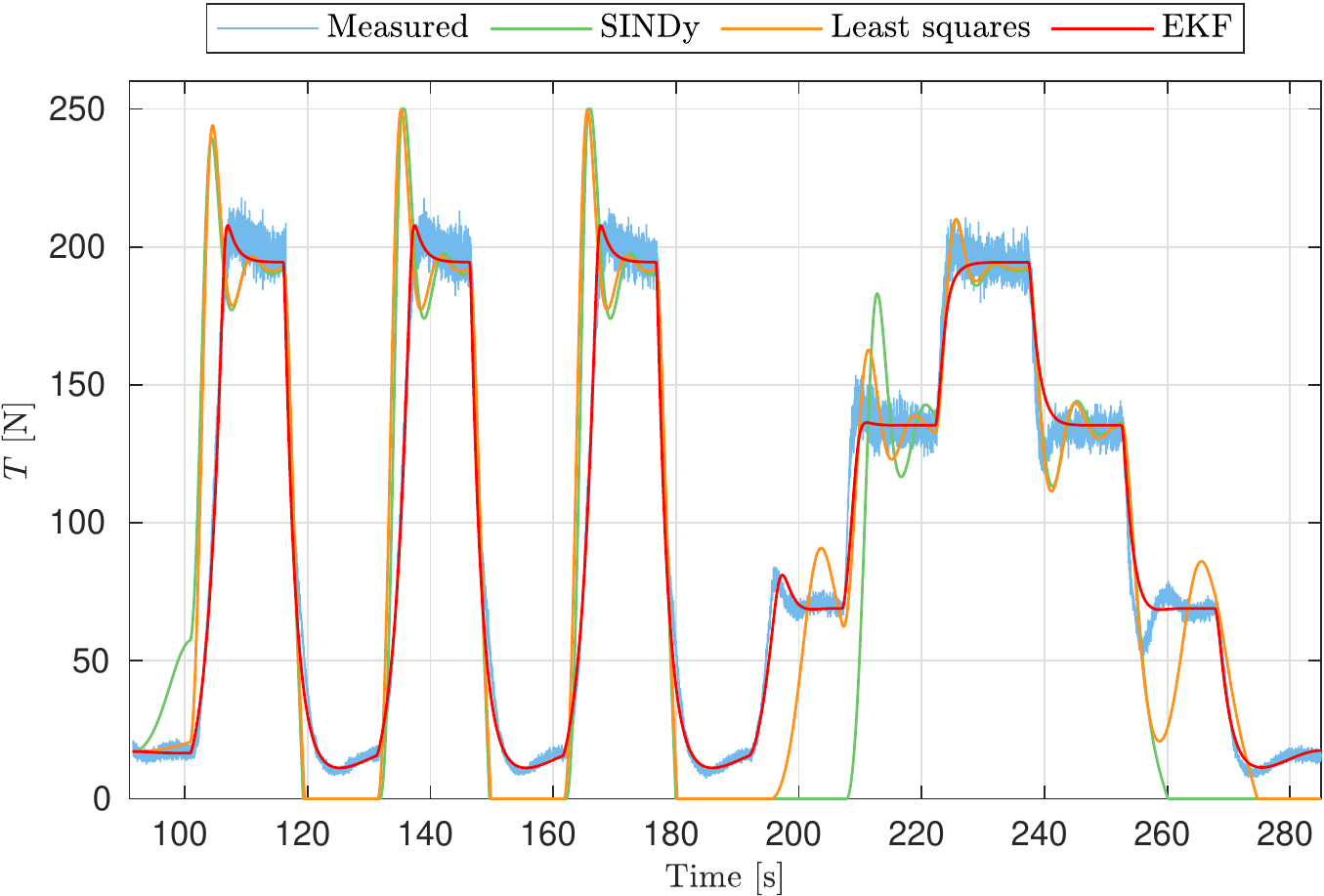}}
 \end{myframe}
 \caption{Results on the calibration dataset (a-c) and validation dataset (b-d).}
 \label{calibration-validation}
\end{figure*}

\subsection{Experimental results of the identification procedure}

In this section, we show the results of the  identification procedures presented in Sec.~\ref{structure} and~\ref{gray-box}. We apply the procedures to identify the model parameters associated with  the P100-RX and the P220-RXi jet engines. When testing the identification procedure based on the EKF,  the variance of the noise measurement, namely $\mathbf{Q}$,  is ${\sim} 7 \ \text{N}^2$ for the P100-RX and ${\sim}9 \ \text{N}^2$ for the P220-RXi. As mentioned above, the dataset used to run the estimations represent ``exciting'' system trajectories, which  guarantee that the regression matrices used in SINDy and batch least squares are full rank, thus making the identification results meaningful.

Fig.~\ref{calibration-validation} shows the experimental data for calibration (left) and validation (right) purposes. Note that all three presented identification procedures are plotted both for the calibration and validation datasets. It is here clear how the EKF based identification procedure outperforms both SINDy and classical iterative least squares. As mentioned previously, this is believed to be the consequence of the intrinsic robustness of the EKF to measurement noise, and to its independence from noncausal filters to estimate higher order thrust derivatives.
Table \ref{tab:turbineParameters} shows the parameters identified by the three identification procedure.
The results on the validation datasets are presented in Fig. \ref{fig:validationP100} and \ref{fig:validationP220}: the EKF method outperforms the other two methods. 
Table \ref{tab:prediction_error} shows the mean absolute error obtained testing the model on 9 different datasets.

\begin{table}[t]
\caption{The identified parameters for both the turbines.}
	\centering
	\begin{tabular}{c|c|c|c|c}
		& \multicolumn{2}{|c|}{EKF} & \multicolumn{2}{|c}{LS} \\
		\midrule
		Parameter & P100-Rx & P220-RXi & P100-Rx & P220-RXi \\
		\midrule
		\rowcolor{gray!15}
		$K_T$     & $-1.460$ & $-0.280$ & $-0.617$ & $0.2027$ \\
		$K_{TT}$  & $-0.059$ & $-0.010$ & $-0.015$ & $-0.003$ \\
		\rowcolor{gray!15}
		$K_D$     & $-2.430$ & $0.5883$ & $-0.737$ & $-0.196$ \\
		$K_{DD}$  & $0.0787$ & $0.0421$ & $-0.002$ & $0.0023$ \\
		\rowcolor{gray!15}
		$K_{TD}$  & $0.1188$ & $0.0058$ & $0.0020$ & $-0.002$ \\
		$B_U$     & $0.4317$ & $0.1874$ & $0.2175$ & $-1.364$ \\
		\rowcolor{gray!15}
		$B_T$     & $0.0116$ & $0.0137$ & $0.0090$ & $-0.002$ \\
		$B_D$     & $-0.026$ & $-0.032$ & $-0.001$ & $0.0067$ \\
		\rowcolor{gray!15}
		$B_{UU}$  & $0.0314$ & $0.0074$ & $0.0078$ & $-0.015$ \\
		$c$       & $-19.92$ & $-7.839$ & $-5.631$ & $20.624$ \\
		\bottomrule
	\end{tabular}
	\label{tab:turbineParameters}
\end{table}

\begin{table}[h]
\caption{Mean absolute error on the validation dataset.}
	\centering
	\begin{tabular}{c|c|c|c}
	    & \multicolumn{3}{|c}{Mean absolute error} \\
		\midrule
		Jet turbine model  & EKF & LS & SINDy  \\
		\midrule
		P100-RX  & $3.94 \ \text{N}$ & $4.40 \ \text{N}$ & $5.11 \ \text{N}$ \\
		P220-RXi & $6.77 \ \text{N}$ & $13.41 \ \text{N}$ & $21.34 \ \text{N}$  \\
		\bottomrule
	\end{tabular}
	\label{tab:prediction_error}
\end{table}
 
\section{CONTROL}
\label{sec:method_control}

In this section, we present strategies to control the model jet engines. We assume that the jet engine  is described by the model~\eqref{turbine_model}, and that the control objective is the asymptotic stabilisation of a desired thrust force $T_\text{d}$.

\subsection{Feedback Linearization control}
In this case, the role of the control input $u$ is to obtain a closed loop dynamics of the form:
\begin{equation}
	\ddot{T} = \ddot{T}_\text{d} + K_\text{p}(T_\text{d} - T) + K_\text{d}(\dot{T}_\text{d} - \dot{T}),
\end{equation}
where $K_\text{p}$ and $K_\text{d}$ are two positive constants. The control law that achieves this closed loop is given by:
\begin{equation}
    \label{feedbackLinear}
	v = \frac{\ddot{T}_\text{d} + K_\text{p}(T_\text{d} - T) + K_\text{d}(\dot{T}_\text{d} - \dot{T}) - f(T,\dot{T})}{g(T,\dot{T})}.
\end{equation}
Then, solving the equation $v = u + B_{UU}u^2$ we obtain the real turbine input $u$. Note that $u \in [25, 100]$ \% by datasheet. In this interval the function $v = u + B_{UU}u^2$ must be positive and monotone in order to take as solution the positive root. 

\begin{figure*}[t]
 \vspace*{0.4cm}
 \centering
 \begin{myframe}{Feedback Linearization VS Sliding Mode control}
 	\vspace{-0.2cm}
 	\hspace{-0.21cm}
 	\subfloat[][{Feedback Linearization control.}\label{fig:fl_accuracy}]
 	{\includegraphics[width=0.502 \linewidth]{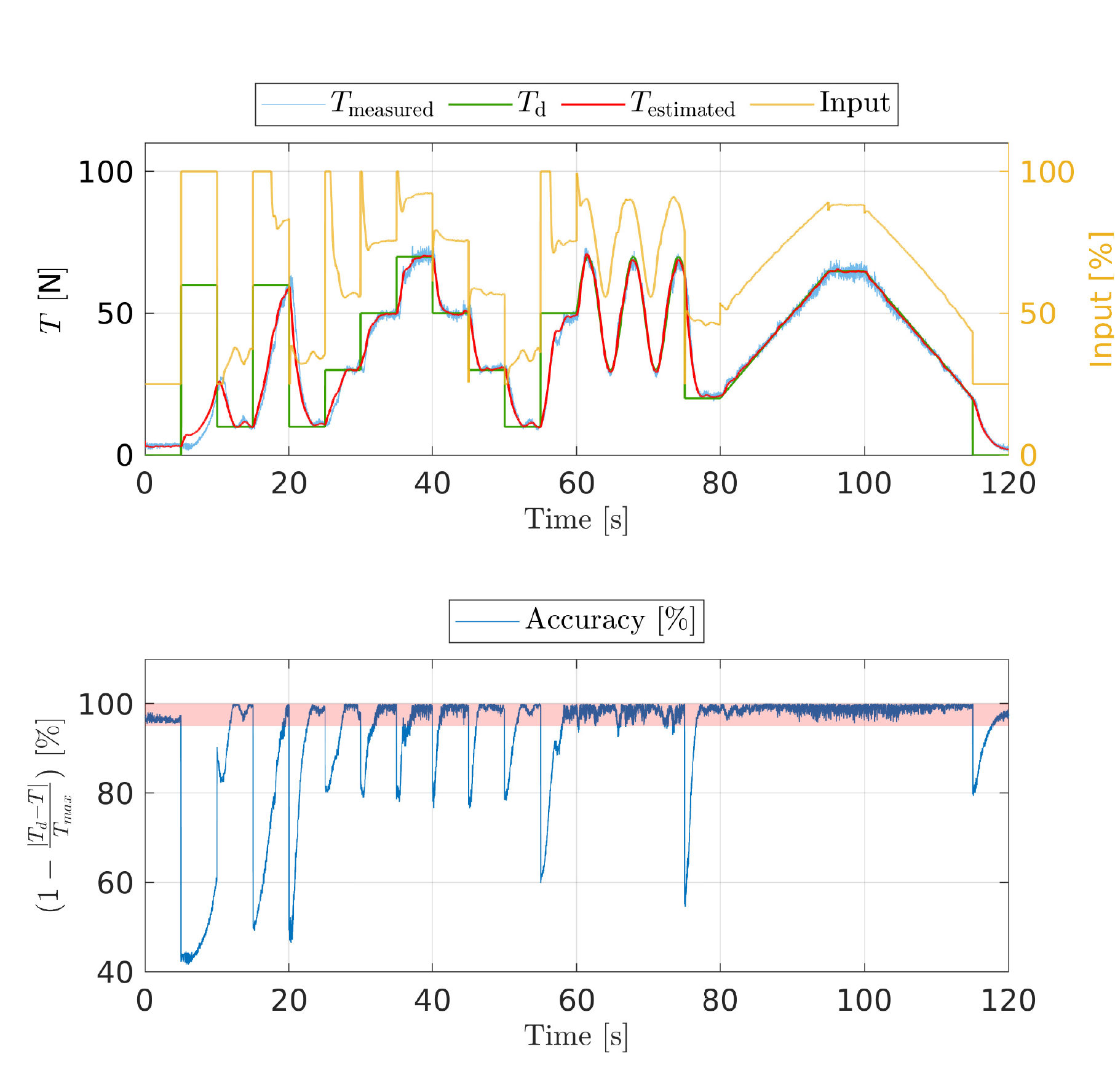}} 
 	\subfloat[][{Sliding Mode control.}\label{fig:sm_accuracy}]
 	{\includegraphics[width=0.502 \linewidth]{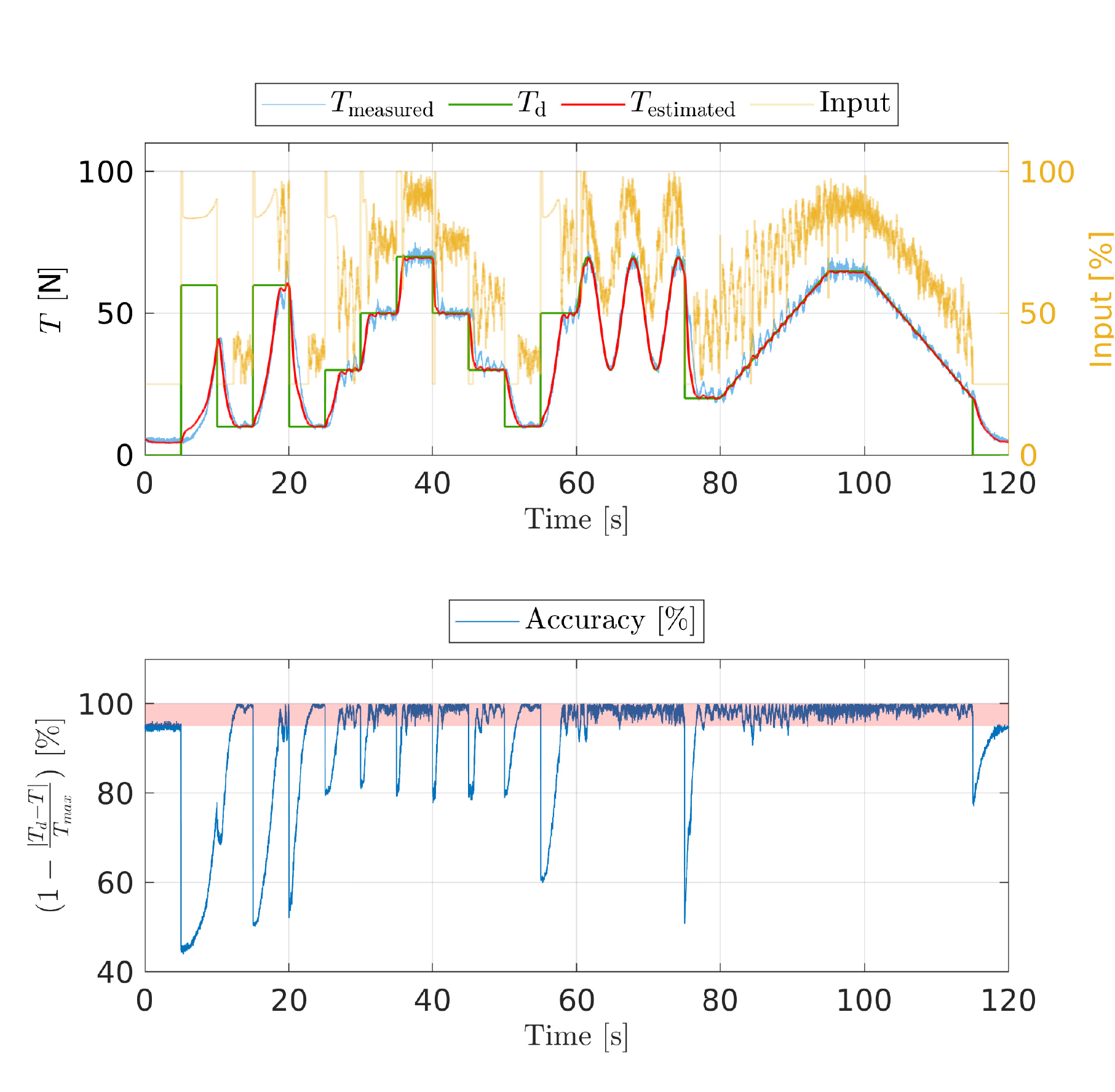}}
 \end{myframe}
 \caption{Feedback Linearization control (a) and  Sliding Mode control (b). The red band identifies the zone in which the error is below the $5$~\%.}
\end{figure*}

\subsection{Sliding mode control}

First, define the following error dynamics:
\begin{equation}
\begin{cases}
		\tilde{T} = T - T_{\text{d}}, \quad
		\dot{\tilde{T}} = \dot{T} - \dot{T}_{\text{d}}, \quad 
		\ddot{\tilde{T}} = \ddot{T} - \ddot{T}_{\text{d}},
\end{cases}\label{errorDynamics}
\end{equation}
which leads to:
\begin{equation}
	\ddot{\tilde{T}} = f(T,\dot{T}) + g(T, \dot{T}) \cdot v(u) - \ddot{T}_\text{d}.
\end{equation}
The \textit{sliding manifold} is then  defined as:
\begin{equation}
	s(\tilde{T}, \dot{\tilde{T}}) = a_1 \tilde{T} + \dot{\tilde{T}},  
\end{equation}
were $a_1$ is a positive constant. To further reduce the aggressiveness of the control law, we can substitute  the switching component of \eqref{sliding_controller}, i.e. the $\text{sgn}(\cdot)$ function, with the continuous approximation $\text{tanh}(\cdot) \in [-1, 1]$.


%
%
%

From \eqref{sliding_controller} and \eqref{errorDynamics}, the control law is then:
\begin{equation}
v = -\frac{a_1 \dot{\tilde{T}} + (f(T, \dot{T}) - \ddot{T}_\text{d})}{g(T, \dot{T})} - \beta \cdot \text{tanh}(K \cdot s),
    \label{slidingModeCtrl}
\end{equation}
where $K$ is an additional positive constant regulating the slope of the $\text{tanh}(\cdot)$ function. We obtain our real turbine input $u$ solving the equation $v = u + B_{UU}u^2$.

\tikzstyle{block} = [draw, fill=black!10, rectangle, 
minimum height=3em, minimum width=6em]
\tikzstyle{sum} = [draw, fill=black!20, circle, node distance=1cm]
\tikzstyle{input} = [coordinate]
\tikzstyle{output} = [coordinate]
\tikzstyle{pinstyle} = [pin edge={to-,thin,black}]

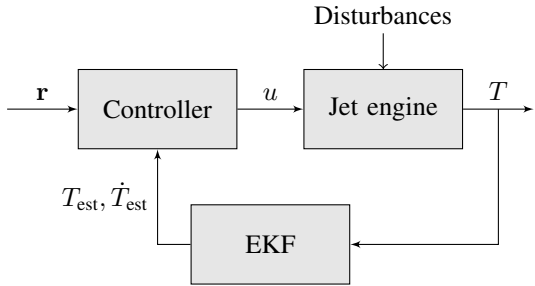
\begin{figure}[t]
	\centering
	\begin{tikzpicture}[auto, node distance=2cm,>=latex']
	\node [input, name=input] {};
	\node [input, name=input2] {};
	\node [block, right of=input] (controller) {Controller};
	\node [block, right of=controller, pin={[pinstyle]above:Disturbances},
	node distance=3cm] (system) {Jet engine};
	\draw [->] (controller) -- node[name=u] {$u$} (system);
	\node [output, right of=system] (output) {};
	\node [block, below of=u] (observer) {EKF};
	
	\draw [draw,->] (input) to node {$\mathbf{r}$} (controller);
	\draw [->] (system) -- node [name=y] {$T$}(output);
	\draw [->] (y) |- (observer);
	\draw [->] (observer) -| node[pos=0.99] {} 
	node [near end] {$T_\text{est}, \dot{T}_\text{est}$} (controller);
	\end{tikzpicture}
	\vspace{0.3cm}
	\caption{The jet engine control architecture, $\mathbf{r} = [T_\text{d}, \dot{T}_\text{d}, \ddot{T}_\text{d}]^\top$.}
	\label{controlArchitecture}
\end{figure}

\subsection{Control architecture and experimental results}

In this section, we show the  performance of the feedback linearization and sliding mode controllers applied to the model jet engine  P100-RX. Let us recall that the only available measurement is the thrust force $T$. We then use an Extended Kalman Filter as state observer  to retrieve the full state $[T,\dot{T}]$. The control architecture is depicted in Fig.~\ref{controlArchitecture}, and the  controller gains are shown in Table~\ref{tab:controllers_gains}. 

Fig. \ref{fig:fl_accuracy} shows the performance obtained by applying the feedback linearization controller~\eqref{feedbackLinear}. Despite the fact that exact feedback linearization methods in general lack robustness, observe that the proposed controller achieves a tracking accuracy of $95$~\% at the steady state. Fig. \ref{fig:sm_accuracy}, instead, depicts  the behavior of the model jet engine when controlled by the sliding mode controller~\eqref{slidingModeCtrl}. This controller generates more aggressive inputs, which cause oscillations of the trajectory along the desired one. Observe, however, that the tracking accuracy of the Sliding Mode controller is also $95$~\% at the steady state, thus leading to similar  performances obtained with the feedback linearization control. 

\begin{table}[t]
\caption{The gains used to evaluate the controllers~\eqref{feedbackLinear} and~\eqref{slidingModeCtrl}.}.
	\centering
	\begin{tabular}{c|c|c|c|c}
	\toprule
	    \multicolumn{3}{c}{Sliding Mode} & \multicolumn{2}{|c}{Feedback Linearization} \\
		\midrule
		$a_1$ & $\beta$ & $K$ & $K_\text{p}$ & $K_\text{d}$  \\
		\midrule
		$20$ & $900$ & $0.15$ & $10$ & $2\sqrt{K_\text{p}}$ \\
		\bottomrule
	\end{tabular}
	\vspace{-0.5cm}
	\label{tab:controllers_gains}
\end{table}

Remark that in both experiments, there are time windows where the accuracy lowers down to about the $40$~\% of the reference signal. Observe, for instance, the first part of the experiments ($0 \ \text{s}-20$~s) that reveals large tracking errors. By looking at the time dependence of the input signals, however, one deduces that low accuracies are mostly due to input saturations when the actual value of the thrust is not close to the desired value.  The controllers cannot deal with this physical limit and the tracking of large steps may then lead to large tracking errors. Moreover, since the minimum input is also bounded (Table~\ref{tab:turbines_specs}) the attainable thrust at idle is not zero. 

Although the Feedback Linearization and Sliding Mode controllers
led to similar tracking accuracies at steady state, let us observe that the tuning of the Sliding Mode one has required careful attention. In fact, the Sliding Mode controller generates a more aggressive control input, and these variations are related to the constant $\beta$. Its tuning, especially for \emph{rapidly} changing desired thrust forces, may not be straightforward in practice, and for these kind of thrust reference we advise to use the feedback linearization controller.

\section{CONCLUSIONS}
\label{sec:conclusion}

This paper presents the first comprehensive approach to the modeling, identification and control  of  model jet engines, with experimental validation of the overall methodology.

Model jet engines are propulsion systems that are still not used in Robotics, and their employment  may pave the way to new robot designs combining several degrees of manipulation and locomotion, as in the case of flying humanoid robots.

This work is based on the main observation that jet engines governing dynamics can be described by a second order nonlinear model. We used  the sparse identification of nonlinear dynamics SINDy to find out the governing model, and then we addressed the identification problem by means of a \textit{gray-box} approach. After system identification, we propose and validate nonlinear controllers for tracking a thrust reference. In particular, we presented feedback linearization and sliding mode controllers. The results show that both  controllers are able to track a given reference, although the Sliding mode control is more sensitive to the gain tuning. 

Let us also underline the importance of having reliable thrust measurements to apply the proposed methodology to robotics platforms. Hence, the development of effective algorithms for thrust estimation is an important step for the proposed approach to be effective.  In the case of the flying humanoid robot iRonCub, however, F/T sensors will be installed in the robot main body, e.g. in the robot arms. Hence, the thrust force will be estimated directly using the robot F/T sensors, a sensor suite similar to that of the testbench presented in the paper. This robot sensor suite may increase the likelihood of the proposed approach to be effective once implemented on the real flying humanoid robot iRonCub.  Future developments will also cover: the design of MPC controllers taking into account  input limitations; adaptive control to compensate for model uncertainties;  the development of a backstepping control technique to use the model jet engines on the iRonCub.


\addtolength{\textheight}{-12cm}

\bibliographystyle{IEEEtran}
\bibliography{IEEEabrv,Biblio}

\begin{thebibliography}{10}
\providecommand{\url}[1]{#1}
\csname url@rmstyle\endcsname
\providecommand{\newblock}{\relax}
\providecommand{\bibinfo}[2]{#2}
\providecommand\BIBentrySTDinterwordspacing{\spaceskip=0pt\relax}
\providecommand\BIBentryALTinterwordstretchfactor{4}
\providecommand\BIBentryALTinterwordspacing{\spaceskip=\fontdimen2\font plus
\BIBentryALTinterwordstretchfactor\fontdimen3\font minus
  \fontdimen4\font\relax}
\providecommand\BIBforeignlanguage[2]{{%
\expandafter\ifx\csname l@#1\endcsname\relax
\typeout{** WARNING: IEEEtran.bst: No hyphenation pattern has been}%
\typeout{** loaded for the language `#1'. Using the pattern for}%
\typeout{** the default language instead.}%
\else
\language=\csname l@#1\endcsname
\fi
#2}}

\bibitem{aerialRobots}
\BIBentryALTinterwordspacing
C.~F. Liew, D.~DeLatte, N.~Takeishi, and T.~Yairi, ``Recent developments in
  aerial robotics: {A} survey and prototypes overview,'' \emph{CoRR}, vol.
  abs/1711.10085, 2017. [Online]. Available:
  \url{http://arxiv.org/abs/1711.10085}
\BIBentrySTDinterwordspacing

\bibitem{shim2005autonomous}
D.~Shim, H.~Chung, H.~J. Kim, and S.~Sastry, ``Autonomous exploration in
  unknown urban environments for unmanned aerial vehicles,'' in \emph{AIAA
  Guidance, Navigation, and Control Conference and Exhibit}, Aug 2005, p. 6478.

\bibitem{DLR_quad}
F.~{Huber}, K.~{Kondak}, K.~{Krieger}, D.~{Sommer}, M.~{Schwarzbach},
  M.~{Laiacker}, I.~{Kossyk}, S.~{Parusel}, S.~{Haddadin}, and
  A.~{Albu-Schäffer}, ``First analysis and experiments in aerial manipulation
  using fully actuated redundant robot arm,'' in \emph{2013 IEEE/RSJ
  International Conference on Intelligent Robots and Systems}, Nov 2013, pp.
  3452--3457.

\bibitem{AirBus_VSR700}
\BIBentryALTinterwordspacing
``Technology breakthrough: Naval group and airbus helicopters responsible for
  building the first demonstrator of a rotary-wing drone for a warship,'' 2018.
  [Online]. Available:
  \url{https://www.airbus.com/newsroom/press-releases/en/2018/01/technology-breakthrough--naval-group-and-airbus-helicopters-resp.html}
\BIBentrySTDinterwordspacing

\bibitem{TechnicalAnalysisVTOLUAV}
S.~Yu, J.~Heo, S.~Jeong, and Y.~Kwon, ``Technical analysis of vtol uav,''
  \emph{Journal of Computer and Communications}, vol.~04, pp. 92--97, Jan 2016.

\bibitem{8299552}
F.~{Ruggiero}, V.~{Lippiello}, and A.~{Ollero}, ``Aerial manipulation: A
  literature review,'' \emph{IEEE Robotics and Automation Letters}, vol.~3,
  no.~3, pp. 1957--1964, July 2018.

\bibitem{6943038}
G.~{Heredia}, A.~E. {Jimenez-Cano}, I.~{Sanchez}, D.~{Llorente}, V.~{Vega},
  J.~{Braga}, J.~A. {Acosta}, and A.~{Ollero}, ``Control of a multirotor
  outdoor aerial manipulator,'' in \emph{2014 IEEE/RSJ International Conference
  on Intelligent Robots and Systems}, Sep. 2014, pp. 3417--3422.

\bibitem{7994965}
M.~Pitonyak and F.~Sahin, ``A novel hexapod robot design with flight
  capability,'' in \emph{2017 12th System of Systems Engineering Conference
  (SoSE)}, June 2017, pp. 1--6.

\bibitem{5334433}
A.~Bozkurt, A.~Lal, and R.~Gilmour, ``Aerial and terrestrial locomotion control
  of lift assisted insect biobots,'' in \emph{2009 Annual International
  Conference of the IEEE Engineering in Medicine and Biology Society}, Sept
  2009, pp. 2058--2061.

\bibitem{6631208}
A.~Kalantari and M.~Spenko, ``Design and experimental validation of hytaq, a
  hybrid terrestrial and aerial quadrotor,'' in \emph{2013 IEEE International
  Conference on Robotics and Automation}, May 2013, pp. 4445--4450.

\bibitem{1748-3190-10-1-016005}
\BIBentryALTinterwordspacing
L.~Daler, S.~Mintchev, C.~Stefanini, and D.~Floreano, ``A bioinspired
  multi-modal flying and walking robot,'' \emph{Bioinspiration \& Biomimetics},
  vol.~10, no.~1, p. 016005, 2015. [Online]. Available:
  \url{http://stacks.iop.org/1748-3190/10/i=1/a=016005}
\BIBentrySTDinterwordspacing

\bibitem{6696526}
L.~Daler, J.~Lecoeur, P.~B. Hählen, and D.~Floreano, ``A flying robot with
  adaptive morphology for multi-modal locomotion,'' in \emph{2013 IEEE/RSJ
  International Conference on Intelligent Robots and Systems}, Nov 2013, pp.
  1361--1366.

\bibitem{LeoCaltech}
\BIBentryALTinterwordspacing
``Caltech building agile humanoid robot by combining legs with thrusters,''
  2019. [Online]. Available:
  \url{https://spectrum.ieee.org/automaton/robotics/humanoids/caltech-building-agile-humanoid-robot-by-combining-legs-with-thrusters}
\BIBentrySTDinterwordspacing

\bibitem{Jet-HR1}
Z.~{Huang}, B.~{Liu}, J.~{Wei}, Q.~{Lin}, J.~{Ota}, and Y.~{Zhang}, ``Jet-hr1:
  Two-dimensional bipedal robot step over large obstacle based on a ducted-fan
  propulsion system,'' in \emph{2017 IEEE-RAS 17th International Conference on
  Humanoid Robotics (Humanoids)}, Nov 2017, pp. 406--411.

\bibitem{Bartolozzi2017iCubTN}
\BIBentryALTinterwordspacing
L.~Natale, C.~Bartolozzi, D.~Pucci, A.~Wykowska, and G.~Metta, ``icub: The
  not-yet-finished story of building a robot child,'' \emph{Science Robotics},
  vol.~2, no.~13, 2017. [Online]. Available:
  \url{https://robotics.sciencemag.org/content/2/13/eaaq1026}
\BIBentrySTDinterwordspacing

\bibitem{pucci2017momentum}
D.~Pucci, S.~Traversaro, and F.~Nori, ``Momentum control of an underactuated
  flying humanoid robot,'' \emph{IEEE Robotics and Automation Letters}, vol.~3,
  no.~1, pp. 195--202, 2017.

\bibitem{nava2018position}
G.~Nava, L.~Fiorio, S.~Traversaro, and D.~Pucci, ``Position and attitude
  control of an underactuated flying humanoid robot,'' in \emph{2018 IEEE-RAS
  18th International Conference on Humanoid Robots (Humanoids)}.\hskip 1em plus
  0.5em minus 0.4em\relax IEEE, 2018, pp. 1--9.

\bibitem{hall2013fluid}
C.~Hall and S.~L. Dixon, \emph{Fluid mechanics and thermodynamics of
  turbomachinery}.\hskip 1em plus 0.5em minus 0.4em\relax
  Butterworth-Heinemann, 2013.

\bibitem{jetEnginesRollsRoice}
{Rolls Royce plc.}, \emph{The jet engine}.\hskip 1em plus 0.5em minus
  0.4em\relax Wiley, 2015.

\bibitem{Brunton3932}
\BIBentryALTinterwordspacing
S.~L. Brunton, J.~L. Proctor, and J.~N. Kutz, ``Discovering governing equations
  from data by sparse identification of nonlinear dynamical systems,''
  \emph{Proceedings of the National Academy of Sciences}, vol. 113, no.~15, pp.
  3932--3937, 2016. [Online]. Available:
  \url{https://www.pnas.org/content/113/15/3932}
\BIBentrySTDinterwordspacing

\bibitem{isidori2013nonlinear}
A.~Isidori, \emph{Nonlinear control systems}.\hskip 1em plus 0.5em minus
  0.4em\relax Springer Science \& Business Media, 2013.

\bibitem{khalil2002nonlinear}
H.~K. Khalil and J.~W. Grizzle, \emph{Nonlinear systems}.\hskip 1em plus 0.5em
  minus 0.4em\relax Prentice hall Upper Saddle River, NJ, 2002, vol.~3.

\bibitem{LIPO_ulvestad2018brief}
A.~Ulvestad, ``A brief review of current lithium ion battery technology and
  potential solid state battery technologies,'' \emph{arXiv preprint
  arXiv:1803.04317}, 2018.

\bibitem{LIPO_endurance}
A.~{Abdilla}, A.~{Richards}, and S.~{Burrow}, ``Power and endurance modelling
  of battery-powered rotorcraft,'' in \emph{2015 IEEE/RSJ International
  Conference on Intelligent Robots and Systems (IROS)}, Sep. 2015, pp.
  675--680.

\bibitem{marketing2007diesel}
{Chevron Products Company}, ``Diesel fuels technical review,'' 2007.

\bibitem{Cat6Fan}
\BIBentryALTinterwordspacing
``The "cat 6" fan,'' 2019. [Online]. Available:
  \url{https://edfdynamax.com/\%22cat-6\%22-plug\%26play-system}
\BIBentrySTDinterwordspacing

\bibitem{P220_ref}
\BIBentryALTinterwordspacing
``Jetcat p220-rxi,'' 2019. [Online]. Available:
  \url{https://www.jetcat.de/en/productdetails/produkte/jetcat/produkte/hobby/Engines/p220-rxi}
\BIBentrySTDinterwordspacing

\bibitem{simon2006optimal}
D.~Simon, \emph{Optimal state estimation: Kalman, H infinity, and nonlinear
  approaches}.\hskip 1em plus 0.5em minus 0.4em\relax John Wiley \& Sons, 2006.

\bibitem{P100_ref}
\BIBentryALTinterwordspacing
``Jetcat p100-rx,'' 2019. [Online]. Available:
  \url{https://www.jetcat.de/en/productdetails/produkte/jetcat/produkte/hobby/Engines/p100_rx}
\BIBentrySTDinterwordspacing

\bibitem{ceseracciu2013middle}
P.~Fitzpatrick, E.~Ceseracciu, D.~Domenichelli, A.~Paikan, G.~Metta, and
  L.~Natale, ``A middle way for robotics middleware,'' \emph{Journal of
  Software Engineering for Robotics}, vol.~5, pp. 42--49, 2014.

\bibitem{RomanoWBI17Journal}
F.~Romano, S.~Traversaro, D.~Pucci, and F.~Nori, ``A whole-body software
  abstraction layer for control design of free-floating mechanical systems,''
  \emph{Journal of Software Engineering for Robotics}, 2017.

\bibitem{ferigo2019generic}
D.~Ferigo, S.~Traversaro, F.~Romano, and D.~Pucci, ``A generic synchronous
  dataflow architecture to rapidly prototype and deploy robot controllers,''
  2019.

\bibitem{schafer2011savitzky}
R.~W. {Schafer}, ``What is a savitzky-golay filter? [lecture notes],''
  \emph{IEEE Signal Processing Magazine}, vol.~28, no.~4, pp. 111--117, July
  2011.

\end{thebibliography}

\end{document}